\documentclass{article}

     \usepackage[final,nonatbib]{neurips_2019}

\usepackage[utf8]{inputenc} 
\usepackage[T1]{fontenc}    
\usepackage{hyperref}       
\usepackage{url}            
\usepackage{booktabs}       
\usepackage{amsfonts}       
\usepackage{amsmath}       
\usepackage{nicefrac}       
\usepackage{microtype}      
\usepackage{cleveref}
\usepackage{graphicx}

\newcommand{\Space}{\Omega}
\newcommand{\Events}{{\cal F}}
\newcommand{\Measure}{P}
\newcommand{\Observables}{\Sigma}
\newcommand{\Rng}[1]{\textrm{range}(#1)}
\newcommand{\Vocab}[1]{\textrm{vocab}(#1)}
\newcommand{\Obs}{M}
\newcommand{\Lang}{{\cal L}}
\newcommand{\Sem}[1]{[\![#1]\!]}
\newcommand{\Prem}{A}
\newcommand{\Conc}{B}
\newcommand{\Interp}{I}
\newcommand{\Ivars}{V}
\newcommand{\QQQ}[1]{P_{#1}}
\newcommand{\Data}{D}
\newcommand{\Pdata}{\QQQ{\Data}}
\newcommand{\UpperFractile}[3]{\textrm{ufr}^{#1}_{#2}{#3}}
\newcommand{\LowerFractile}[3]{\textrm{lfr}^{#1}_{#2}{#3}}
\newcommand{\AtomicInterp}{{\sc AtomicInterp}}
\newcommand{\ConjInterp}{{\sc ConjInterp}}
\newcommand{\True}{\textrm{true}}

\newtheorem{example}{Example}
\newtheorem{definition}{Definition}

\title{Bayesian Interpolants as Explanations for Neural Inferences}

\author{%
  Kenneth L.~McMillan \\
  Microsoft Research
}

\begin{document}

\maketitle

\begin{abstract}
  The notion of Craig interpolant, used as a form of explanation in
  automated reasoning, is adapted from logical inference to
  statistical inference and used to explain inferences made by neural
  networks. The method produces explanations that are at the same time
  concise, understandable and precise. 
\end{abstract}

\section{Introduction}
Statistical machine learning models, most notably deep neural
networks, have recently proven capable of accomplishing a remarkable
variety of tasks, such as image classification, speech recognition,
and visual question answering, with a degree of precision formerly
only achievable by humans. There has been increasing concern, however,
about the opacity of these models. That is, while the models may
produce accurate predictions, they do not produce explanations of
these predictions that can be understood and validated by humans.  As
noted by O'Neil~\cite{WMD}, this opacity can have substantially
negative social consequences. From a purely technical point of view,
it creates an obstacle to improving models and methods.

This paper addresses the question of what it means for a model to
produce a meaningful explanation of a statistical prediction or inference.  We
take as our starting point a notion of explanation used in automated
reasoning. In Boolean satisfiability (SAT) solving~\cite{DBLP:series/faia/2009-185} and model checking~\cite{DBLP:reference/mc/2018},
one typically reasons from particular cases to the general case.  For
example, in SAT solving, we may consider a partial assignment of truth
values to Boolean variables. In model checking, we might consider a
particular control path of a program. To generalize from a particular
case, we need an \emph{explanation} of why a property holds in
that case. An explanation has two key characteristics: it must be a
\emph{simple} proof of the property, and it must be expressed in a
\emph{form} that can be generalized. The notion of simplicity is a
bias: it prevents over-fitting the explanation to the particular
case. The formal or syntactic requirement depends on the kind of generalization we
wish to perform.

In automated reasoning, the explanation is often expressed as a
\emph{Craig interpolant}. Given a logical premise $A$ that implies a conclusion $B$, an
\emph{interpolant} for this implication is an intermediate predicate
$I$, such that:
\begin{itemize}
\item $A$ implies $I$, 
\item $I$ implies $B$, and
\item  $I$ is expressed only using variables that are in common to $A$ and $B$.
\end{itemize}
Notice that the first two requirements are semantic, while the last is
syntactic. This is crucial to the interpretation of interpolants as
explanations.  From an intuitive point of view, for $A$ to explain a
fact to $B$, it must speak $B$'s language, abstracting away concepts irrelevant
to $B$. It is this abstraction that affords generalization. The
notion of interpolation as explanation was proposed from a
philosophical point of view by Hintikka and Halonen~\cite{doi:10.1086/392742}.
From a practical point of view, McMillan showed that interpolation
could be used to derive facts about intermediate states in specific
program executions. These facts could be generalized to an inductive invariant
proving the general case~\cite{DBLP:conf/cav/McMillan03}. In retrospect, the
conflict clauses~\cite{DBLP:journals/tc/Marques-SilvaS99} used in SAT solvers
can also be seen as interpolants. In these applications,
the key properties of an interpolant are that it should be
\emph{simple}, in order to avoid over-fitting, and it should use the
right vocabulary, in order to abstract away irrelevant concepts.

How, then can we transfer this notion of explanation of a
\emph{logical} inference to some notion of explanation of a
\emph{statistical} inference? In this paper, we will explore the idea
of using a kind of naive Bayesian ``proof''. An inference in this
proof will be of the form $P(B\ |\ A) \geq \alpha$ where $\alpha$ is a
desired degree of certainty.  We will maintain the notion of finding
an intermediate fact in the proof that uses the ``right''
vocabulary. In the statistical setting, however, our premise $A$ and
conclusion $B$ need not have any variables in common. This is because
the premise and conclusion are connected by an underlying probability
distribution. Thus, instead of using the common vocabulary, we will
simply choose some set of variables $V$ that we consider to be
causally intermediate between premise $A$ and conclusion~$B$. We then
define a \emph{(na\"ive) Bayesian interpolant} to be a predicate~$I$ over
the intermediate variables~$V$ such that $P(I\ |\ A) \geq \alpha$ and
$P(B\ |\ I) \geq \alpha$. That is, if we observe~$A$, then $I$ is
probably true, and if we observe $I$, then $B$ is probably true. This
is a ``na\"ive'' proof because it implies that
$P(A\ | \ B) \geq \alpha^2$, but only under the unwarranted assumption
that $A$ and $B$ are independent given $I$.

We will explore the application of Bayesian interpolants as
explanations for inferences made by artificial neural networks
(ANN's). In this case, the premise~$A$ is an input presented to the
network, the conclusion~$B$ is the prediction made by the network, and
the vocabulary~$V$ represents (typically) the activation of the
network at some hidden layer. We will observe that,
despite possibly unjustified assumptions, interpolation can produce
explanations that are both understandable to humans, and remarkably 
precise. 

The paper is organized as follows. In \cref{sec:interp}, we formalize
the notion of Bayesian interpolant. In \cref{sec:learn}, we define
inference of interpolants as a multi-objective optimization problem,
and introduce a simple framework for learning interpolants from
data. In \cref{sec:mnist}, we apply the approach to a convolutional
neural model for the MNIST digit recognition problem, interpreting the
results and comparing them to existing methods.

\section{Bayesian interpolants}
\label{sec:interp}

Let $\Observables$ be a set of \emph{observables}. Each observable $v$
has a \emph{range}, denoted $\Rng{v}$. An \emph{observation} is a map
that takes each observable to an element of its range. Now let
$(\Space,\Events,\Measure)$ be a probability space, where
$\Space$ is a set of observations, $\Events$ is a $\sigma$-algebra
over $\Space$ (the measurable sets) and $\Measure$ is the probability
measure, assigning probabilities to sets in $\Events$.

\begin{example}
  We are given an ANN that classifies images of hand-written digits
  with $n$ layers, the first being the input layer and the last the
  output layer. The observable $v_{i,j}$ is the activation of the
  $j$-th unit in the $i$-th layer, while observable $w$ is the digit
  prediction. We have $\Rng{v_{i,j}}=\mathbb{R}$ and $\Rng{w}=\{0\ldots 9\}$.
  The observables in layers $i>1$ are
  functionally determined by the input layer.  That is, for $i>1$ and any
  observation $\Obs\in\Space$, we have
  $\Obs(v_{i,j}) = f_{i,j}(\Obs(v_{1,1}\ldots v_{1,k}))$ and
  $M(w) = \textrm{arg max}_{j=0\ldots 9} M(v_{n,j})$. 
\end{example}

Let $\Lang$ be a language of \emph{formulas}. The
\emph{vocabulary} of a formula $\phi\in \Lang$, denoted
$\Vocab{\phi}$, is a subset of the observables $\Observables$.  If
$\Obs$ is an observation and $\phi$ is a formula, we write
$\Obs \models \phi$ to indicate that $\Obs$ is a model of $\phi$ (that
is, $\phi$ is true in $\Obs$).  We write $\Sem{\phi}$ for the set of models
of $\phi$. We require that the truth value of a formula depends only
on its vocabulary, that is, if two observations $\Obs$ and $\Obs'$
agree on the observables in $\Vocab{\phi}$, then $\Obs \models \phi$
if and only if $\Obs' \models \phi$. We further require that $\Lang$ is
closed under conjunction. That is, if $\phi$ and $\phi'$ are formulas,
then so is $\phi \wedge \phi'$, and moreover:
\begin{itemize}
\item $\Vocab{\phi \wedge \phi'} = \Vocab{\phi} \cup \Vocab{\phi'}$
\item $\Obs \models \phi\wedge\phi'$ iff $\Obs \models \phi$ and
  $\Obs \models \phi'$.
\end{itemize}
Finally, we require that, for all $\phi\in\Lang$, the models of $\phi$ form a
measurable set, that is, $\Sem{\phi}\in\Events$. In the sequel, we will abuse notation by writing
simply $\phi$ for $\Sem{\phi}$ when the intention is clear. For
example, we will write $P(\phi)$ for the probability that $\phi$ is
true, which is $\Measure(\Sem{\phi})$.

\begin{example}
  For the ANN example, let $\Lang$ be the set of finite conjunctions
  of atomic predicates of the form $v \leq c$, $v \geq c$ and $v= c$,
  where $v$ is an observable and $c$ is a numeric constant, for example
  $v_{2,42}\leq 0.2 \wedge v_{2,76} \geq 3.5$.
\end{example}

\begin{definition}
Suppose that $\Prem$ and $\Conc$ are formulas in $\Lang$, $\Ivars$
is a set of observables, and $\alpha$ is a probability in $[0,1]$. We
say that a $(\Ivars,\alpha)$-interpolant is formula $\Interp$ such
that $\Vocab{\Interp} \subseteq \Ivars$ and:
\begin{align}
  P(\Interp\ |\ \Prem) &\geq \alpha \label{eq:interpleft}  \\
  P(\Conc\ |\ \Interp) &\geq \alpha \label{eq:interpright} 
\end{align}
provided the above conditional probabilities are well-defined.
\end{definition}

\begin{example}
  We wish to explain why the ANN classifies a particular image $x$ as
  digit~7.  In this case, our premise $\Prem$ is
  $\wedge_{i,j} v_{1,(i,k)} = x_{i,j}$ (stating that the input image
  is $x$) and our conclusion $B$ is $w = 7$. We seek a fact about
  layer 2 that explains the prediction~7. Thus, we let
  $\Ivars=\{v_{2,i}\}$. After a search, we discover a
  $(\Ivars,0.95)$-interpolant:
  $v_{2,42}\leq 0.2 \wedge v_{2,76} \geq 3.5$. This fact about a pair
  of layer~2 units holds for image $x$ and predicts that the network
  will output~7 with precision~0.95. We consider this fact an explanation
  because it is simple and expressed over variables that are causally
  intermediate between input and output.
  
\end{example}

An additional figure of merit for an interpolant is its \emph{recall},
which we define as $P(I\ |\ B)$. The recall is the fraction of
positive occurrences of the conclusion~$B$ that are predicted by the
interpolant. The recall gives us a measure of how likely the
interpolant is to be applicable in cases where the premise $A$ is
false. While high recall is not required for an explanation, we will
observe that a minimum value is needed for statistical significance.

\begin{definition}
A
$(\Ivars,\alpha,\gamma)$-interpolant is a $(\Ivars,\alpha)$-interpolant that also
satisfies:
\begin{equation}
  P(\Interp\ |\ \Conc) \geq \gamma  \label{eq:interprecall}
\end{equation}
\end{definition}

\section{Learning interpolants}
\label{sec:learn}

In the case of learned models, the underlying probability space is
hypothetical. In place of the distribution, we are given a data set
that is presumed to be an i.i.d. sample from the distribution. Thus, we
cannot solve \crefrange{eq:interpleft}{eq:interprecall} exactly, but
instead must treat these constraints as hypotheses to be tested. In
this section, we will introduce a machine learning approach to
discover interpolants. The method uses a training data set for
hypothesis formation and an independent test data set for confirmation.

As might be expected, there is a trade-off among precision, recall and
simplicity. For example, we may increase simplicity at the expense of
recall, or recall at the expense of precision. The learning approach
has hyper-parameters to control this trade-off. Moreover, recall, training set size and variance are related. That is,
if the number of samples satisfying~$\Interp$ is low, then the
confidence in the statistical test of \cref{eq:interpright} is
correspondingly low. Thus we will find that generalization is likely
to fail, in the sense that precision on the test set is lost.

A \emph{dataset} is a finite indexed family of observations.
A dataset $\Data$ induces a probability measure distribution $\Pdata$ such that $\Pdata(\phi)$ is
the frequency of satisfaction of $\phi$ in the dataset, that is:
\begin{equation}
  \label{eq:pd}
  \Pdata(\phi) = \frac{|\{\Data_i\ |\ \Data_i \models \phi\}|}{|\Data|}
\end{equation}
We will write $\Data\ | \ \phi$ for the sub-family $\{\Data_i\ | \ \Data_i \models \phi\}$.

Interpolation can be seen as a multi-objective constrained
optimization problem. That is, given a dataset $\Data$, we would like
to solve \crefrange{eq:interpleft}{eq:interprecall} for $P = \Pdata$
while maximizing $\alpha$, $\gamma$ and minimizing the syntactic complexity
of~$I$.

It is tempting to try to apply standard learning approaches,
\emph{e.g.}, gradient boosted trees~\cite{friedman2001} or random forests~\cite{DBLP:journals/ml/Breiman01},
to solve this optimization problem. However, while these methods allow
a degree of trade-off between recall and precision, they are not
well-suited to explore the Pareto front. We cannot, for example, use
them to solve for the most precise classifier of a certain complexity
having a fixed recall. 


Instead, we introduce here a technique that solves the multi-objective
optimization problem for atomic predicates by exhaustive search. It
then uses a simple boosting technique that conjoins new predicates in
order to improve precision at the expense of recall, until a target
precision is reached. Thus, we attempt to minimize complexity and
maximize recall, for a fixed precision,

\subsection{Atomic Bayesian interpolants}

An atomic predicate is one without any logical connectives, such as
conjunction or if-then-else. As an example, in decision tree learning,
the atomic predicates label the decision nodes and are usually of the
form $v \leq c$ or $v \geq c$, where $v$ is an observable and $c$ is a
constant. We will call these upper (respectively lower) bound
predicates.  For these predicates, it is practical to learn optimal
interpolants by exhaustive search.

Given fixed values of $\alpha$ and $\gamma$, we can find bound predicates
that satisfy \cref{eq:interpleft} and \cref{eq:interprecall} by computing
\emph{fractiles}. We define the upper fractile $\UpperFractile{p}{v}{\phi}$, where $p$ is a probability, $v$ an observable,
and $\phi$ a formula,
as the least $c$ such that $P(v\leq c\ |\ \phi) \geq p$. Similarly, the lower fractile $\LowerFractile{p}{v}{\phi}$ is 
the greatest $c$ such that $P(v\geq c\ |\ \phi) \geq p$. For $P = \Pdata$, we can compute fractiles
in $O(N\textrm{log}N)$ time, where $N$ is the number of data points, by simply sorting $\Data\ |\ \phi$ with respect to $v$.

\Cref{fig:atomicinterp} shows a procedure \AtomicInterp\ that computes
an atomic interpolant~$I$ for given values of $\alpha$ and $\gamma$,
optimizing the precision $P(\Conc\ |\ I)$. The set $L$ contains, for
each observable~$v$, the strongest lower bound predicate satisfying
\cref{eq:interpleft} and \cref{eq:interprecall} (thus, the one with
minimal recall). The set $U$ contains the corresponding upper bound
predicates. Among these, the procedure chooses the one with the
maximal precision. Minimizing recall (subject to the constraints) is
not guaranteed to find the interpolant with \emph{maximal}
precision. However, it is a good heuristic, since in general precision
increases with decreasing recall.

\begin{figure}
  \begin{tabbing}
    mm\=mm\=mm\=mm\=mm\=mm\=\kill
    \> \textbf{procedure} \AtomicInterp($\Prem$,$\Conc$,$\Data$,$\Ivars$,$\alpha$,$\gamma$)\\
    \> \> \textbf{requires} $\Pdata(\Prem) > 0$ and $\Pdata(\Conc) > 0$ \\
    \> \> \textbf{with} $P = \Pdata$:\\
    \> \> \> $L \leftarrow \{v\geq c \ |\ v\in\Ivars, c = \textrm{min}(\LowerFractile{\alpha}{v}{A},\LowerFractile{\gamma}{v}{B})\}$\\
    \> \> \> $U \leftarrow \{v\leq c \ |\ v\in\Ivars, c = \textrm{max}(\UpperFractile{\alpha}{v}{A},\UpperFractile{\gamma}{v}{B})\}$\\
    \> \> \> $I \leftarrow \textrm{arg max}_{\phi \in L \cup U} P(B\ | \phi)$\\
    \> \> \> \textbf{return} $I$\\
  \end{tabbing}

   \caption{Computing an atomic interpolant, optimizing precision.}
   \label{fig:atomicinterp}
\end{figure}

\begin{figure}
  \begin{center}
  \includegraphics[width=4in]{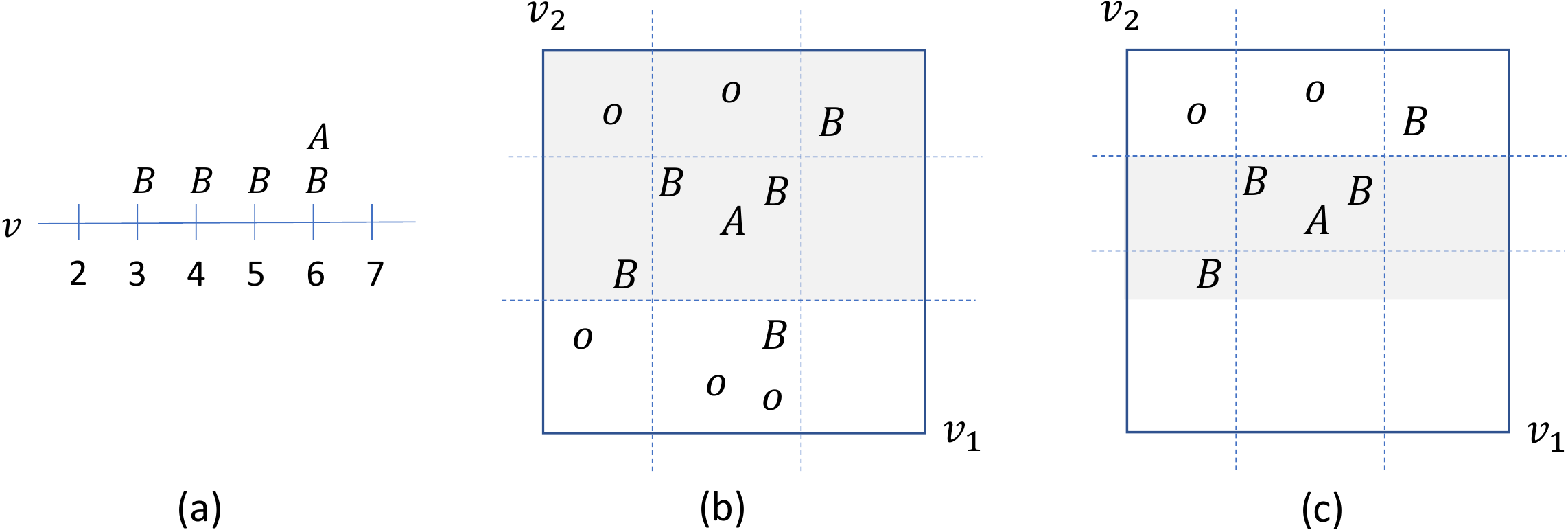}
\end{center}
\caption{Interpolant learning examples. The symbol $B$ represents points where $B$ is true,
  $A$ represents points where \emph{both} $A$ and $B$ are true, and $o$ represents points where
neither are true.}
  \label{fig:interpexamps}
\end{figure}

\begin{example}
  Suppose $V = \{v\}$, $\alpha=0.95$, $\gamma=0.75$ and $D$ is
  distributed as in \cref{fig:interpexamps}(a).  The relevant fractiles
  are $\LowerFractile{\alpha}{v}{A}= 6$,
  $\LowerFractile{\gamma}{v}{B}=4$, $\UpperFractile{\alpha}{v}{A}= 6$,
  $\UpperFractile{\gamma}{v}{B}= 5$. This gives us $L = \{v \geq 4\}$
  and $U = \{ v \leq 6\}$. We have $P(B\ |\  v \geq 4) = 3/4$ while
  $P(B\ |\ v \leq 6) = 4/5$. Thus, we return $v \leq 6$.
\end{example}

\subsection{Conjunctive Bayesian interpolants}

A single atomic predicate is unlikely to provide sufficient precision
to act as a useful explanation. The procedure of \cref{fig:conjinterp}
boosts the precision of the interpolant by conjoining predicates until
a precision goal is reached.  At each step, the recall
constraint~$\gamma$ is reduced by a factor $\mu$ to allow the
interpolant to be strengthened. Moreover, we keep only the data points satisfying
the current interpolant~$I$, to focus on eliminating the remaining false positives.
The procedure terminates when the
precision goal $\alpha$ is achieved, or the complexity of the
interpolant reaches a maximum~$\kappa$.  As a measure of the
complexity of $\phi$, we can use the number of bound predicates, which
we denote $|\phi|$.  


\begin{figure}
  \begin{tabbing}
    mm\=mm\=mm\=mm\=mm\=mm\=\kill
    \> \textbf{procedure} \ConjInterp($\Prem$,$\Conc$,$\Data$,$\Ivars$,$\alpha$,$\gamma$,$\mu$,$\kappa$)\\
    \> \> \textbf{requires} $\Pdata(\Prem) > 0$ and $\Pdata(\Conc) > 0$ \\
    \> \> $I \leftarrow \True$\\
    \> \> \textbf{while} $\Pdata(B\ |\ I) < \alpha$ and $|I| < \kappa$:\\
    \> \> \> $I \leftarrow I \wedge \textrm{\AtomicInterp}(\Prem,\Conc,\Data,\Ivars,\alpha,\gamma)$\\
    \> \> \> $\Data \leftarrow \Data\ |\ I$\\
    \> \> \> $\gamma \leftarrow \mu\gamma$\\
    \> \> \textbf{return} $I$\\
  \end{tabbing}

   \caption{Computing an conjunctive interpolant, optimizing recall and complexity.}
   \label{fig:conjinterp}
\end{figure}

\begin{example}
  Suppose $V=\{v_1,v_2\}$, $\alpha=0.95$, $\gamma=0.8$ and $\mu=0.9$, and
  suppose the dataset $D$ is distributed as in
  \cref{fig:interpexamps}(b). The dotted lines show the $\gamma$-th
  fractiles. The interpolant $I$ at the first step is the shaded
  region.  It has a precision of $5/7$ and a recall of
  $5/6$. \Cref{fig:interpexamps}(b) shows the result of boosting the
  precision by conjoining a predicate. We consider only the points
  satisfying $I$.  The resulting precision is $1$, and the recall is
  $4/6$. Since the precision is greater than~$\alpha$, the algorithm
  terminates.
  
\end{example}

If both~$\alpha$ and $\gamma$ are close to one, then
interpolation becomes essentially transfer learning. That is, we are
attempting to learn a new predictor as a function of a
representation~$\Ivars$ learned by the ANN. In this case, however, we
would necessarily sacrifice simplicity, and thus explanatory power. By
relaxing the requirement for recall, we gain in simplicity, and are
thus able to form useful explanations. For this purpose, $\gamma$ and $\mu$
should lie in a ``sweet spot'', where both complexity and variance
are low.

The procedure described above is just an example of a method for
learning interpolants. Conjunctions of bound predicates are a natural
choice, since they are easy to compute and to understand, and they
allow us to boost precision by sacrificing recall. Other sorts of
atomic predicates are possible, however, for example, linear
constraints. One could also generalize conjunctions to, say, decision
trees. An important difference between interpolation and common
learning problems is that, in seeking an explanation for a particular
inference, high recall is not required but some lower bound on recall
is needed. This means that existing learning approaches may need to be
modified to learn interpolants.  Nonetheless, one could attempt to
explore the space of interpolants by using existing loss-based
approaches and simply scaling down the loss associated with false
negatives. If this is effective, it would open up a wide range of
possible approaches.

\section{Example: MNIST digit recognition task}
\label{sec:mnist}

In this section, we experiment with using Bayesian interpolants as
explanations for inferences made by ANN's. As an example, we
use the MNIST hand-written digit database~\cite{LeCun1998}. This consists of
labeled $28\times 28$-pixel gray-scale images of hand-written digits,
and has been widely used as a benchmark for both learning and explanation
methods.

The ANN model we use has five layers: 1) the input layer, 2) a linear
convolutional layer with $3\times 3$ kernel and 28 channels
($26\times 26\times 28 = 18,928$ units in total), 3) a $2\times 2$ max-pooling
layer ($13\times 13\times 28 = 4,732$ units in total), 4) a fully-connected
layer of 128 units with RELU activation, and 5) a fully-connected
output layer of 10 units with soft-max activation. This model is
implemented in the Keras framework~\cite{Keras} and trained on the
60,000-image MNIST training sample using the Adam optimizer with
sparse categorical cross-entropy loss function. It achieved 98.2\%
accuracy over the 10,000-image MNIST test set.

We begin by computing interpolants at layer~3, the max-pooling layer,
because this layer has 1/4 of the units of the convolutional layer,
but is still localizable in the image (each unit corresponds to
$4\times 4$ region of pixels). For each experiment, we choose an input
image $x$ whose output category (\emph{i.e.}, the output unit with highest activation)
is $y$. We wish to produce an explanation for this inference, so our premise $\Prem$
is $v_1 = x$ (which we take as an abbreviation for $\wedge_j v_{1,j}=x_j$)
and our conclusion~$\Conc$ is $w=y$. To ensure that $\Pdata(A) > 0$, we must
include image $x$ in the dataset. 

\begin{figure}
  \begin{center}
      \includegraphics[width=5.0in]{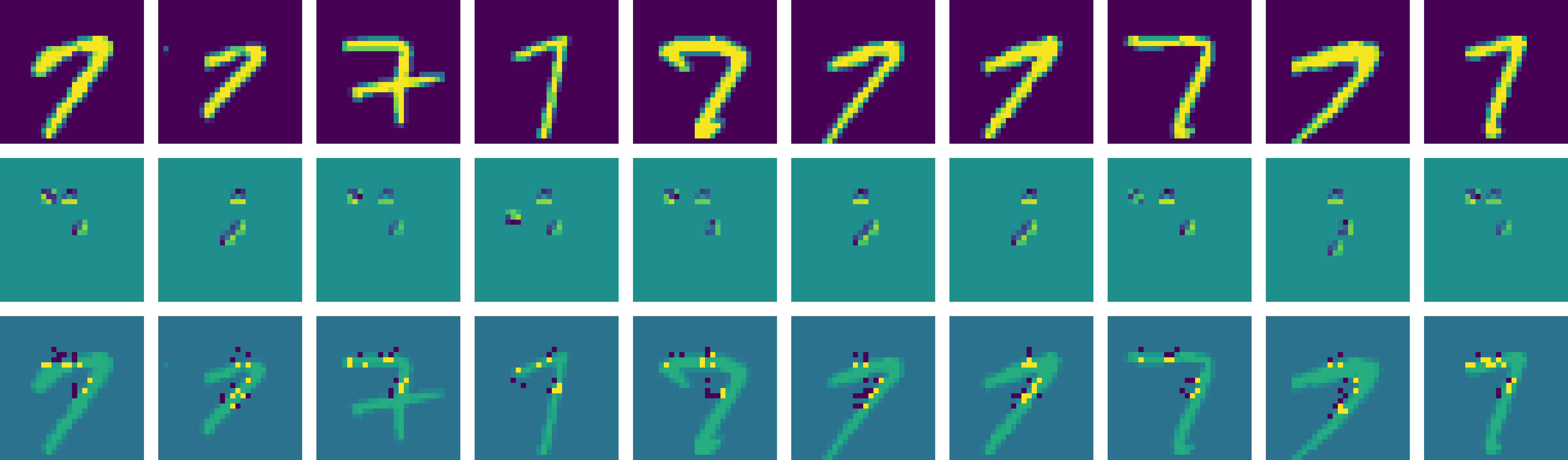}

      \vspace{0.1in}
      
      \includegraphics[width=5.0in]{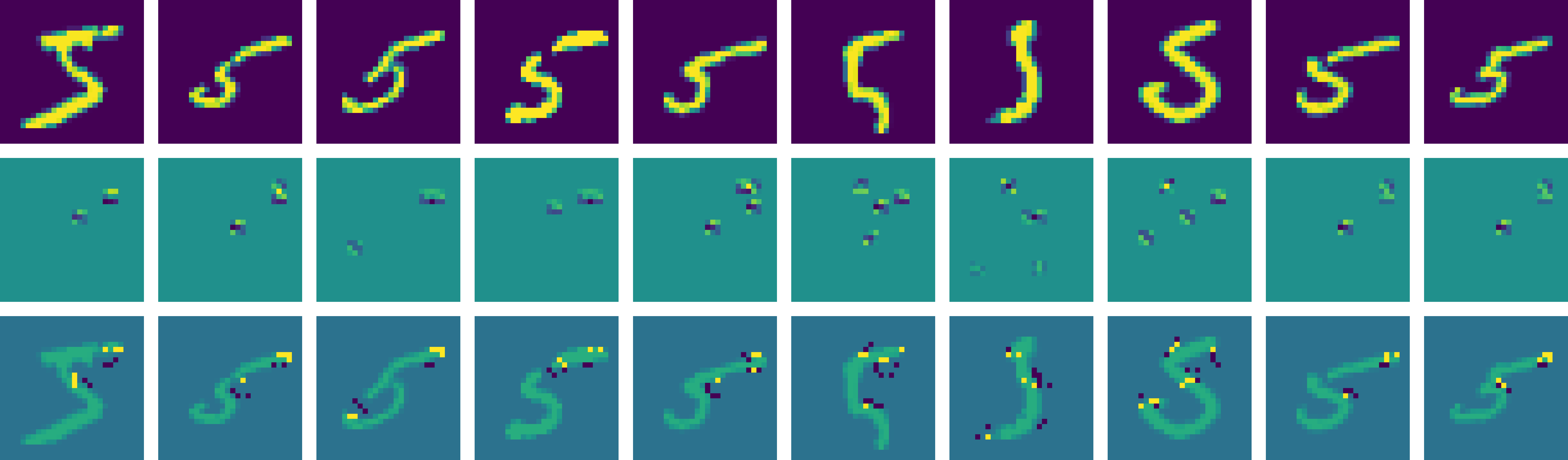}

      \vspace{0.1in}

      \includegraphics[width=5.0in]{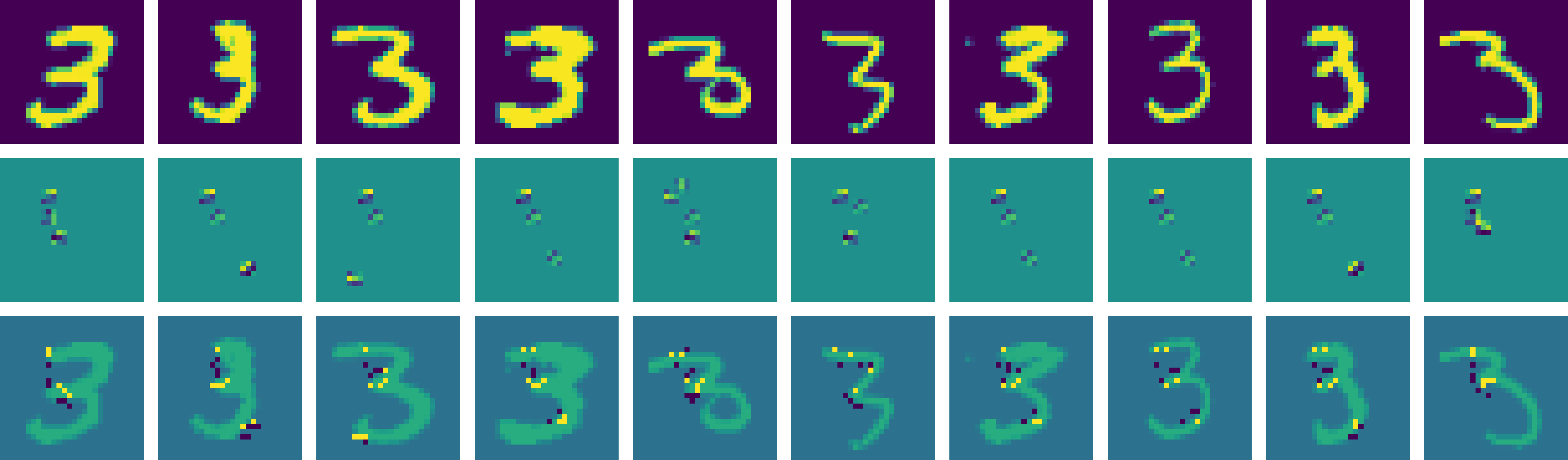}
  \end{center}
  \caption{Visualization of interpolants learned for MNIST model inferences.}
  \label{fig:mnist1}
\end{figure}

We compute interpolants using procedure \ConjInterp\ where $\Data$ is
the first 20,000 images in the training set. \Cref{fig:mnist1} shows
interpolants computed for the first 10 images in the training set
classified by the model as digits 7, 5 and 3, with $\alpha=0.98$,
$\gamma=0.55$ and $\mu=0.9$ (the digits were chosen in advance of
seeing the data, to avoid cherry-picking). For each case, the input
images are shown in the top row.  The middle row images are a
superposition of the kernels of the units used in the interpolant. Due
to pooling, each $3\times 3$ kernel shown is actually applied over a
region of $4\times 4$ pixels.  For the digit seven, we can see that the
typical number of units used in the interpolant is three out of 4,732,
while for the more complex figure three, typically 3--4 units are needed.
The average complexity of the interpolants for digit seven is 3.2
conjuncts, the average precision is 0.982, and the average recall is
0.123. That is, just three out of 4,732
units are needed to predict the outcome with 98\% precision.

\begin{figure}
  \begin{center}
      \includegraphics[width=2.6in]{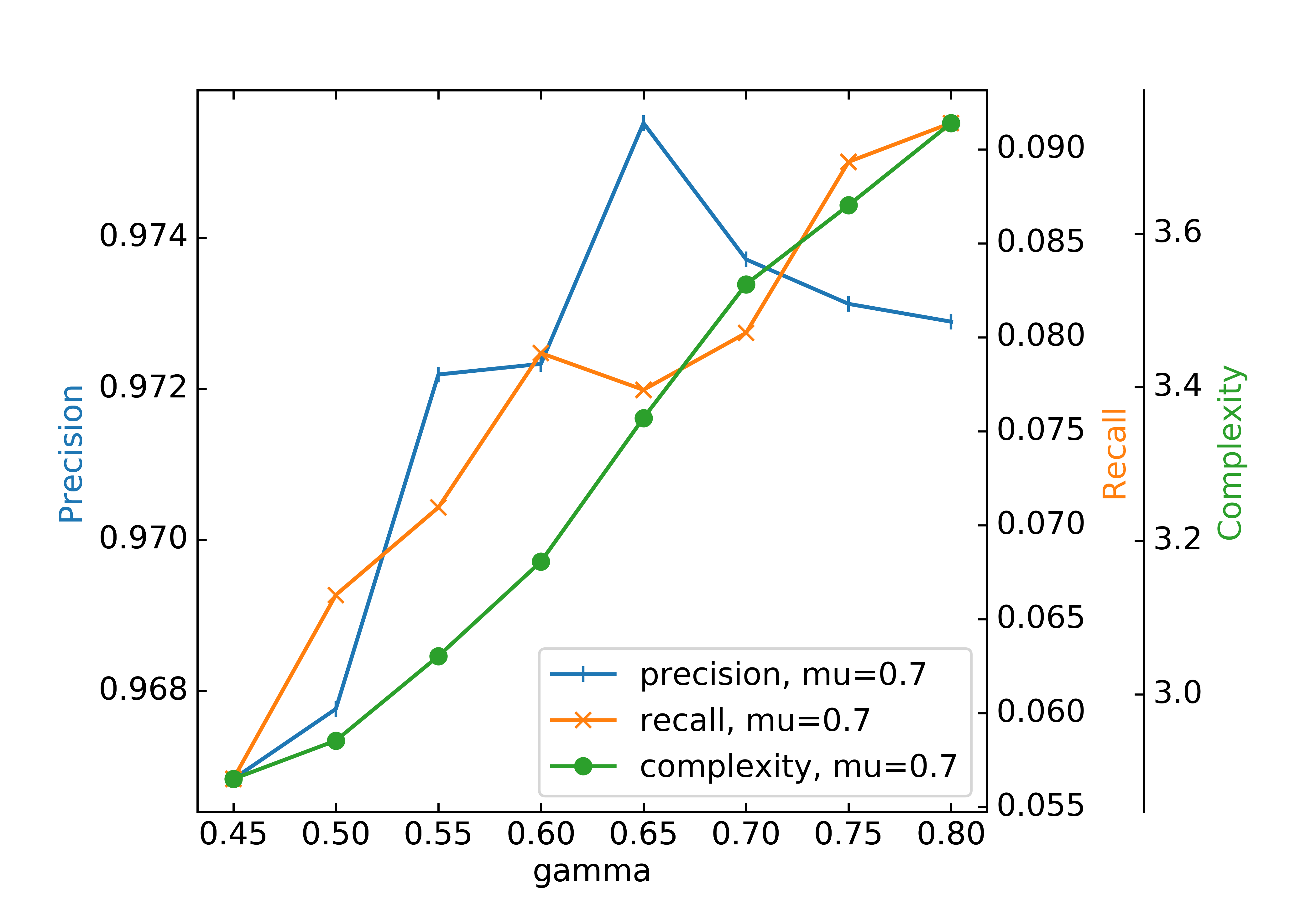}  \includegraphics[width=2.6in]{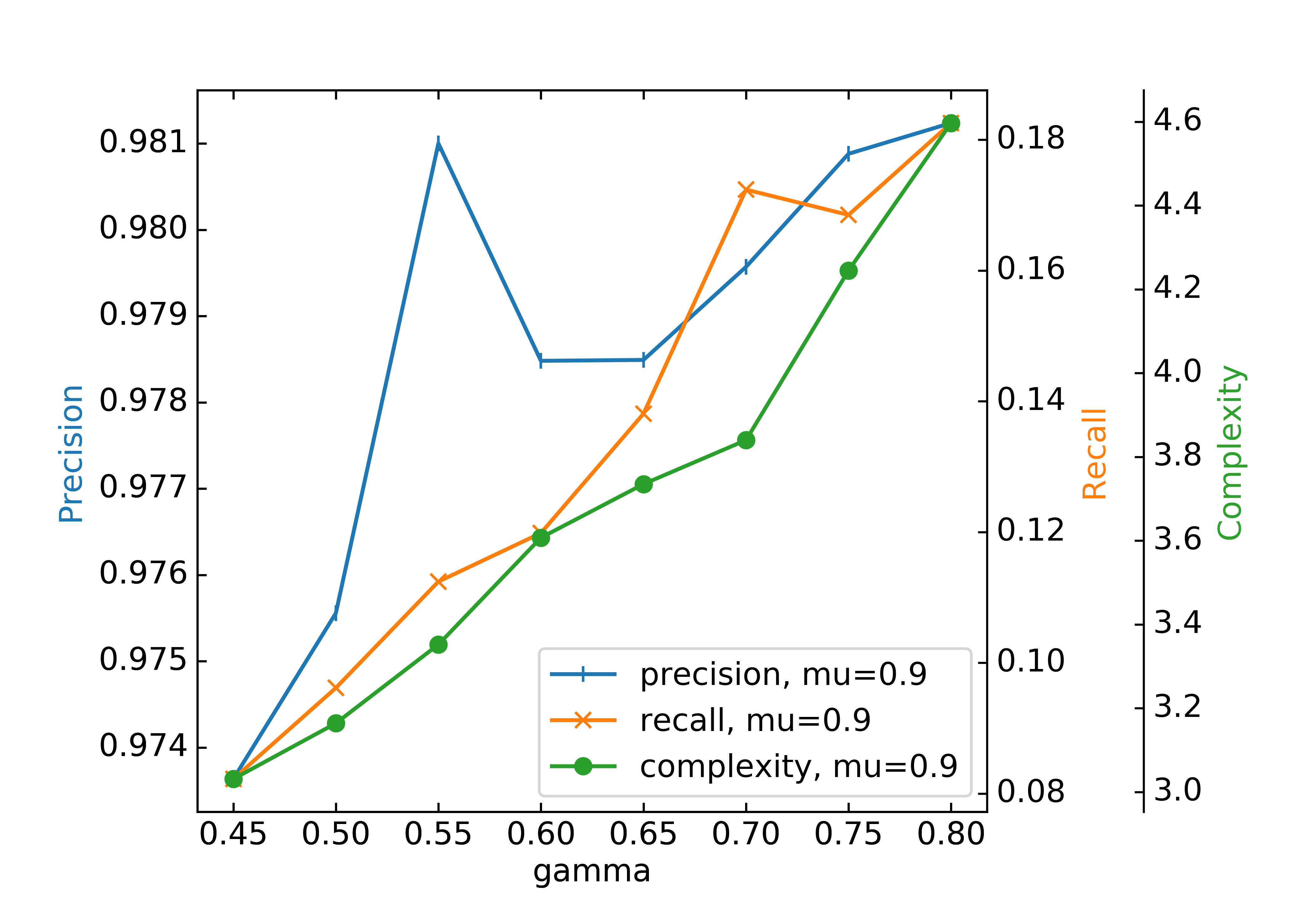}
  \end{center}
  \caption{Precision, recall, and complexity of layer-3 interpolants.}
  \label{fig:sweep}
\end{figure}

In \cref{fig:sweep}, we see plots of the average precision,
recall and interpolant complexity obtained, as a function of the
hyper-parameters $\gamma$ and $\mu$, keeping $\alpha$ fixed at
$0.98$. Complexity is measured as the number of atomic predicates. The
figures are averaged over 1000 input images, selected as the first 100
images classified by the ANN as digit~$n$, for $n = 0\ldots 9$. To
ensure that the interpolants generalize, precision and recall are
calculated over the \emph{test} dataset, not the training dataset. As
we increase $\gamma$ (the lower bound on the recall of atomic
interpolants) the recall of the interpolant increases. At
the same time, the average number of conjuncts in the interpolant
increases, reflecting the fact that more boosting steps are needed to
achieve the target precision for a higher recall. Notably, precision
on the test set increases slightly with recall (note that the precision
scale in the plots has a very narrow range). This may seem
counter-intuitive, since the opposite relationship typically holds.
Notice, however, that with lower recall, the interpolant $I$ is
true on fewer samples in the training set. Optimizing the precision
$P(B\ |\ I)$ over a small set of samples can lead to over-fitting of
the interpolant, and thus to reduced precision on the test
set. Therefore, while we do not require high recall for purposes of
explanation, we need to set some lower bound on recall in order for
the interpolants to generalize well. The peaks in precision may be caused by
the fact that precision on the \emph{training} set drops off slightly with
increasing recall and complexity. Thus there is a ``sweet spot'' balancing training
precision and generalization.

\Cref{fig:sweep} clearly shows a trade-off of simplicity of the
interpolant \emph{vs.} precision.  To be a useful explanation, the
interpolant must be both simple and highly predictive, thus we must
make this trade-off carefully. In the MNIST application, a sweet spot
occurs at around 3.3 conjuncts, with precision around
$0.98$, which we can achieve with parameter values $\gamma=0.55$
and $\mu=0.9$.



\subsection{Sequence interpolants}

While the layer-3 interpolants give a sense of the general regions
of the image that are important for predicting the output, the kernels
themselves are difficult to interpret. For example, in some cases,
complex kernels that appear to be edge or line detectors are actually
being used in dark regions of the image.  This is interesting, as the ANN
may be opportunistically using features for
purposes for which they are not optimally tuned.

To explain the activation of these units by a particular image, we can
apply interpolation a second time. That is, we learn an interpolant at
the input layer (layer 1) that predicts the layer-3 interpolant.  In
effect, we are constructing a na\"ive Bayesian proof in three steps:
the image predicts a fact about the input layer, which in turn
predicts a fact about an internal layer, which in turn predicts the
output. Such a multi-stage proof could be called a \emph{sequence
  interpolant}.


For the layer-1 interpolant we can take advantage of the convolutional
structure of the network to introduce a regularization. That is, we
know that each convolutional unit is causally dependent on only a
small region of the input pixels. Thus, for each conjunct $I_{3,i}$ of
the layer~3 interpolant $I_3$, we learn a layer-1 interpolant
$I_{1,i}$ over just the $4\times 4$ patch of the input image on which
the corresponding layer-3 unit depends. Our layer-1 interpolant $I_1$
is the conjunction $\wedge_i I_{1,i}$.  In this way, we exploit our knowledge
that certain observables are relevant in order to
introduce bias, and thus hopefully better generalization than we would
obtain by directly learning a layer-1 interpolant that predicts the
output. We will see that this strategy can be quite effective.

In \cref{fig:mnist1}, the third row of images for each digit shows a
representation of the atomic predicts in the layer-1 interpolant
thus obtained, with the input image faintly superposed. Like the
layer-3 interpolants, the layer-1 interpolants are computed with
$\alpha=0.98$, $\gamma=0.55$ and $\mu=0.9$. A dark magenta pixel
indicates an upper bound predicate, while a bright yellow pixel indicates a
lower bound predicate. The light green pixels are not part of the
interpolant. These interpolants are much easier to interpret than the
kernels of the layer-3 interpolant. For example, in the case of
digit seven, it is clear that the most important predictive feature is
typically the corner in the upper right of the seven. Above an to the left
of the two strokes, a dark region is required. For digit five,
the top horizontal stroke is most prominent (with exceptions for two oddly
formed digits). Notice that the interpolant for the oddly shaped
figure five in column~7 is still consistent with a more typically
shaped five. For digit three, the three horizontal segments are
most prominently used. To get a sense of how strong these interpolants are as
predictions, try to draw a realistic figure of a different digit
(centered in the image) that passes through the bright points without
touching the dark points.

How precise are these interpolants in predicting the ANN's output?
Each interpolant $I_{1,i}$ predicates the corresponding atomic
predicate $I_{3,i}$ with precision of about $\alpha=0.98$, and the
conjunction of the $I_{3,i}$ in turn predicts the outcome with about
the same precision.  Using the na\"ive assumption that these
probabilities are independent, we would estimate $I_1$ predicts the
output with a precision of about $\alpha^4$ or $0.92$. To evaluate the
actual obtained precision, we used as a test set the 40,000 held-out
images from the training set. We need a large test set because the
average recall of $I_1$ is only 0.005. Even with large test set, 51\%
of the 1000 interpolants were true on zero images in the test set,
thus the precision could not be estimated, as the denominator was
zero. On the remaining 49\% of interpolants, the average precision was
0.996. The cumulative precision of all predictions made by the
interpolants was also 0.996 (that is, over all 1000 interpolants, the
false positive rate was 0.004).

This anomalously high precision requires explanation. Previously, we
observed that low recall led to reduced precision, due to over-fitting
to a small sample. In this case, however, there is no
over-fitting because the $I_{1,i}$ are learned independently and over different features of
the input space. Like a random forest, this is an example of stochastic discrimination~\cite{DBLP:journals/amai/Kleinberg90}.
Each
$I_{1,i}$ has by itself a good precision and moderate recall over the test set in
predicting $I_{3,i}$. By forming the conjunction, we in effect boost
this precision over both the training and test sets, at the expense of
recall. Notice that, unlike a random forest, the weak predictors in our interpolant
have been selected to predict features of another model that explain its output in a particular case.
This is very different from random selection (\emph{i.e.}, bagging) but has a similar effect
of improving generalization. 

What is still puzzling, however, is that $I_1$ actually
predicts the output more strongly than $I_3$. One way to interpret
this is that, because of correlations in the input space, $I_1$
predicts \emph{stronger} bounds at layer~3 than those represented by
the conjuncts of $I_3$.  Thus, because the output depends causally on
layer~3, it is more strongly predicted by $I_1$ than by $I_3$. This
implies that we could reasonably trade some precision for simplicity
in~$I_1$ by reducing the parameter~$\alpha$. 

It is remarkable to observe that the small fraction of pixels used in
the interpolants of \cref{fig:mnist1} (on average 12.4 out of 784) can
predict the ANN output with such high precision. One way to look at this is that we
obtained a combination of simplicity and generality by using the
causal structure of the ANN as a regularizer.

We also note that the interpolants are not expensive to compute.  The
average time to compute both interpolants for the MNIST model is 2.8s on
an 8-core, 3.6GHz Intel Xeon processor, using the numpy library in
Python.

\subsection{Comparison to other explanation techniques}

\begin{figure}
  \begin{center}
  \includegraphics[width=5.0in]{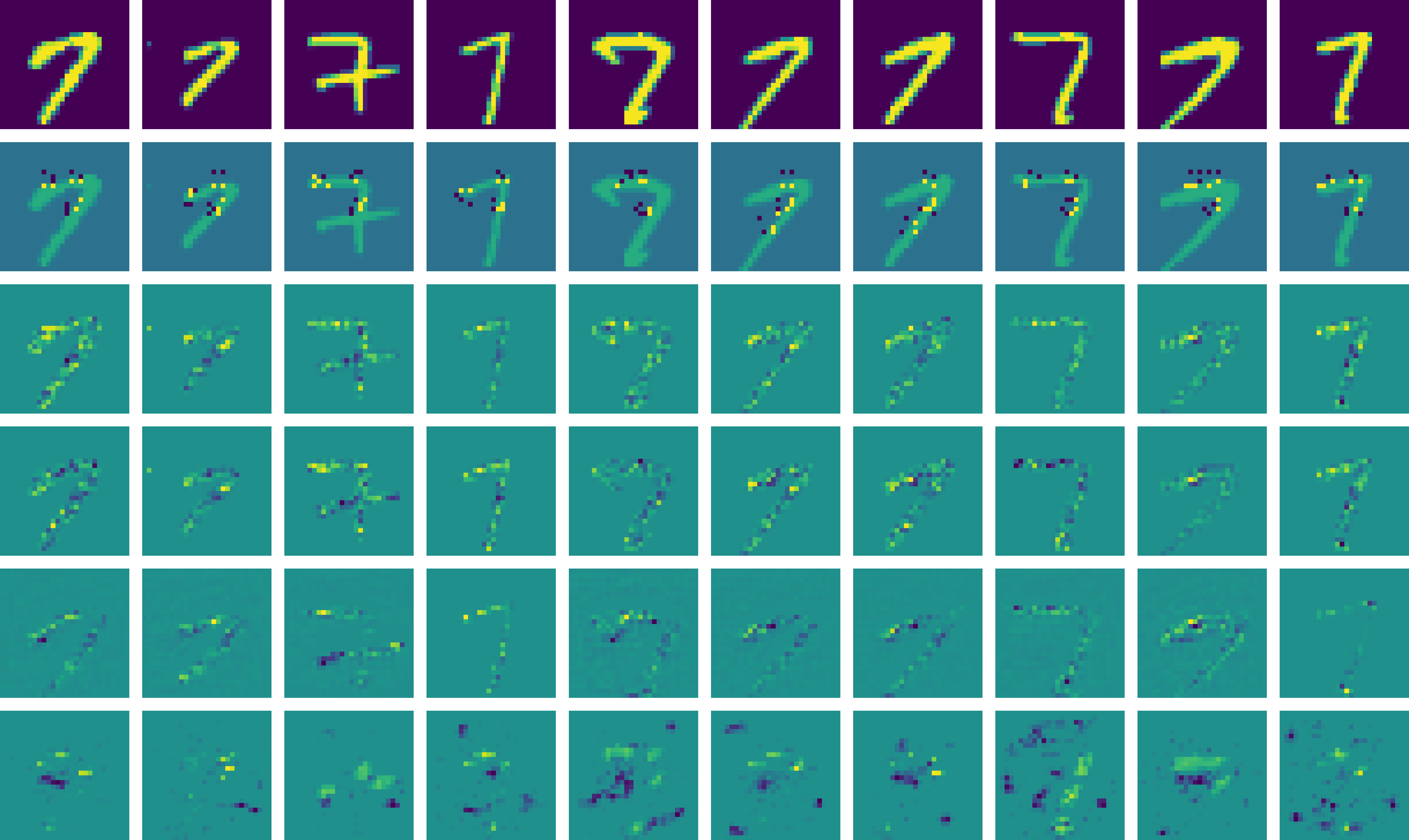}
  \end{center}
  \caption{Comparison of interpolation with gradient-based
    methods. Rows: 1) input image, 2) interpolant 3) Integrated Gradients
    4) DeepLift 5) Grad-CAM 6) CEM.}
  \label{fig:compare7}
\end{figure}

Many techniques have been explored for producing
human-understandable explanations of neural inferences, especially for
image classification tasks. Many use the same gradient
computation that is used for training the network to identify input
units of high sensitivity, relative to a baseline
input. \Cref{fig:compare7} compares of the layer-1
interpolants to the results of three gradient-based methods:
Integrated Gradients ~\cite{DBLP:conf/icml/SundararajanTY17}, DeepLift~\cite{DBLP:journals/corr/ShrikumarGK17}, and Guided
Grad-CAM~\cite{DBLP:conf/iccv/SelvarajuCDVPB17} (all as implemented in the Captum tool~\cite{Captum})
as well as the Contrastive Explanation Method (CEM) which is based on
adversarial generation~\cite{DBLP:conf/nips/DhurandharCLTTS18}, as implemented by its authors~\cite{CEM}. 

Results are shown for the first 10 images categorized as digit seven by
the network. In the gradient-based methods, bright yellow pixels are
attributed as positively influencing the digit seven output, while dark
magenta pixels are negatively influencing. Because the all-zero image is used
as the baseline, these methods do not indicate
relevance of background pixels. For CEM, the bright pixels represent
a subset of the image's foreground pixels that is classified as seven.
The dark pixels represent background pixels that, if activated, would
cause the image to be classified as not seven.

We observe that the Integrated Gradients and DeepLift methods do not
localize the inference. That is, almost all foreground pixels
are assigned positive or negative influence. The Grad-CAM method does
somewhat localize some images.  All of the methods attribute negative
influence to many foreground pixels.  None offers any clear
explanation of the inference. On the other hand, the interpolants are
specific, and they point to a key feature in the recognition of a
digit seven: the sharp angle in the upper right. Moreover, the
interpolants show the influence of background pixels in recognizing
the angle, while the gradient methods do not consider the
background. Most importantly, the interpolants are \emph{predictive}:
they predict the inference with an average precision of 99.6\% over
the test set. The gradient-based explanations only give a general
notion that certain regions of the image may be more relevant than
others.

The CEM method does succeed in some cases in localizing the
explanation (columns 1--3,6,7).  However, in these cases the
explanation is not precise, as it is possible to draw a figure other
that seven that is consistent with the images (for example a two in
cols. 1,2,6,7, or a four in col. 3). On the other hand, the
explanations that appear to be precise are not simple. Thus the method
does not make a good trade-off between simplicity and precision.

\section{Conclusion}

In automated reasoning, we often seek explanations in terms of simple
proofs that can be generalized. This idea is used to learn from
failures in SAT solving, to refine abstractions in model checking, and
to generalize proofs about loop-free programs into inductive proofs
about looping programs.  In this paper, we have explored a statistical
analog of this concept of explanation, with the purpose of giving
explanations of inferences made by statistical models such as
ANN's. The form of the explanation is a Bayesian interpolant: a chain
of statistical inferences leading from a premise to a conclusion. The
structure of this chain is informed by the causal structure of the
model, which acts as a regularizer or bias, allowing us to construct a
simple and yet highly precise explanation.

It is important to understand that a Bayesian interpolant is a
statistical explanation, not a causal explanation. In principle, it is
possible that the observables that are used to express an interpolant
are not even logically connected to the output, but are merely
correlated with it. If our concern is in gaining confidence in the
model's inference, and not in \emph{how} the model arrives at it, then
this is not an issue.  However, interpolants do tell us
\emph{something} about how the ANN works. The existence of very simple
interpolants with high precision tells us the representations used by
the ANN are highly redundant.
As in automated deduction, interpolants may point to important
features in the ANN, or they may suggest more predictive features
than the ones actually used. They might also allow us to compress
the network or compile it into a more efficiently computable form.

Because Bayesian explanations are quantitative predictions, they may
provide a way to address the problem of fairness in statistical
models. That is, suppose we have a model that is trained on a dataset
that reflects bias on the part of human decision makers, or
societal bias that impacts outcomes. How can we determine whether a
given inference made by the model is fair or rational? If the
inference comes with a Bayesian explanation that depends only on
variables that we consider to be a rational basis for decisions, and
yet at the same time is highly predictive of the outcome, then we may
consider the inference to be acceptable. On the other hand, if the
explanation depends on variables that we consider irrelevant or
irrational, then we may reject it. Moreover, a simple and predictive
explanation may act as a guide in the case of an undesired outcome, and
reduce the opacity of the model to those affected by its inferences.


\bibliographystyle{plain}
\bibliography{bib}

\end{document}